\documentclass[11pt,a4paper]{article}
\usepackage[hyperref]{LM_QG_wjzhou}
\usepackage{times}
\usepackage{latexsym}

\usepackage{url}

\usepackage{booktabs}
\usepackage{graphicx}
\usepackage{amsmath}
\usepackage{amssymb}
\usepackage{multirow}
\usepackage{verbatim}
\usepackage{color}
\usepackage{caption}
\usepackage{float}

\aclfinalcopy 

\title{Multi-Task Learning with Language Modeling for Question Generation}

\author{Wenjie Zhou, Minghua Zhang, Yunfang Wu\thanks {Corresponding author.} \\
Key Laboratory of Computational Linguistics, Ministry of Education \\
School of Electronics Engineering and Computer Science, Peking University, Beijing, China \\
{\tt \{wjzhou013,zhangmh,wuyf\}@pku.edu.cn}
}

\date{}

\begin{document}
\maketitle
\begin{abstract}
This paper explores the task of answer-aware questions generation. Based on the attention-based pointer generator model, we propose to incorporate an auxiliary task of language modeling to help question generation in a hierarchical multi-task learning structure. Our joint-learning model enables the encoder to learn a better representation of the input sequence, which will guide the decoder to generate more coherent
and fluent questions. On both SQuAD and MARCO datasets, our multi-task learning model boosts the performance, achieving state-of-the-art results. Moreover, human evaluation further proves the high quality of our generated questions. 
\end{abstract}

\section{Introduction}
Question generation (QG) receives increasing interests in recent years due to its benefits to several real applications: (1) QG can aid in the development of annotated questions to boost the question answering systems \cite{DBLP:conf/emnlp/DuanTCZ17,DBLP:journals/corr/TangDQZ17}; (2) QG enables the dialogue systems to ask questions which make it more proactive \cite{DBLP:journals/jzusc/ShumHL18,Colby:1975:AP:542894}; (3) QG can help to generate questions for reading comprehension texts in the education field. In this paper, we focus on answer-aware QG. Giving a sentence and an answer span as input, we want to generate a question whose response is the answer.

Previous work on QG was mainly tackled by two approaches: the rule-based approach and neural-based approach. The neural-based approach receives a booming development due to the release of large-scale reading comprehension datasets like SQuAD \cite{DBLP:conf/emnlp/RajpurkarZLL16} and MARCO \cite{DBLP:conf/nips/NguyenRSGTMD16}. Most of the neural approaches on QG employ the encoder-decoder framework, which incorporate attention mechanism to pay more attention to the informative part and copy mode to copy some tokens from the input text \cite{DBLP:conf/acl/DuSC17,DBLP:conf/nlpcc/ZhouYWTBZ17,DBLP:conf/naacl/SongWHZG18,DBLP:conf/acl/SubramanianWYZT18,DBLP:conf/emnlp/ZhaoNDK18,DBLP:conf/emnlp/SunLLHMW18}. To make better use of answer information, \citet{DBLP:conf/naacl/SongWHZG18} leverage multi-perspective matching, and \citet{DBLP:conf/emnlp/SunLLHMW18} propose a position-aware model that aims at putting more emphasis on the answer-surrounded context words. \citet{DBLP:conf/emnlp/ZhaoNDK18} aggregate paragraph-level information to help QG. Another line of work is to deal with question answering and question generation as dual tasks \cite{DBLP:journals/corr/TangDQZ17,DBLP:conf/emnlp/DuanTCZ17}. Some other works try to generate questions from a text without answers as input \cite{DBLP:conf/acl/SubramanianWYZT18,DBLP:conf/emnlp/DuC17}. Although some progress has been made, there is still much room for improvement for QG. 

Multi-task learning is an effective way to improve model expressiveness via related tasks by introducing more data and fruitful semantic information to the model \cite{Caruana:1998:ML:296635.296645}. Many works in NLP have adopted multi-task learning and prove its effectiveness on textual entailment \cite{DBLP:conf/emnlp/HashimotoXTS17}, keyphrase generation \cite{DBLP:conf/emnlp/YeW18} and document summarization \cite{DBLP:conf/acl/BansalPG18}. To the best of our knowledge, no work attempts to employ multi-task learning for question generation. Although language modeling has been applied to multi-task learning for classification tasks, they are different from our generation task. 

In this work, we propose to incorporate language modeling as an auxiliary task to help QG via multi-task learning. We adopt the pointer-generator \cite{DBLP:conf/acl/SeeLM17}  reinforced with features as the baseline model, which yields state-of-the-art result \cite{DBLP:conf/emnlp/SunLLHMW18}. The language modeling task is to predict the next word and the previous word whose input is a plain text without relying on any annotation. The two tasks are then combined with a hierarchical structure, where the low-level language modeling encourages our representation to learn richer language features that will help the high-level network to generate better expressive questions. 

We conduct extensive experiments on two reading comprehension datasets: SQuAD and MARCO. We experiment with different settings to prove the efficacy of our multi-task learning model: with/without language modeling and with/without features. Experimental results show that the language modeling consistently yields obvious performance gain over baselines, for all evaluation metrics, including \emph{BLEU}, \emph{perplexity} and \emph{distinct}. Our full model outperforms the existing state-of-the-art results on both datasets, achieving a high BLEU-4 score of 16.23 on SQuAD and 20.88 on MARCO,         respectively. We also conduct human evaluation, and our generated questions get higher scores on all three metrics, including $matching$, $fluency$ and $relevance$. 

\begin{table*}[!htp]
\centering
\resizebox{\textwidth}{!}{
\begin{tabular}{l@{\
}|c@{\quad}c@{\quad}c@{\quad}c@{\quad}c@{\quad}c@{\quad}c@{\quad}c@{\quad}}
\toprule
\textbf{Dataset} & \multicolumn{4}{c}{\textbf{SQuAD}} & \multicolumn{4}{c}{\textbf{MARCO}} \\
\textbf{Model} & \textbf{BLEU-1} & \textbf{BLEU-2} & \textbf{BLEU-3} & \textbf{BLEU-4} & \textbf{BLEU-1} & \textbf{BLEU-2} & \textbf{BLEU-3} & \textbf{BLEU-4} \\
\midrule
\midrule
NQG++ \cite{DBLP:conf/nlpcc/ZhouYWTBZ17} & - & - & - & 13.29  & - & - & - & - \\
matching strategy \cite{DBLP:conf/naacl/SongWHZG18} & - & - & - & 13.91  & - & - & - & - \\
Maxout Pointer (sentence) \cite{DBLP:conf/emnlp/ZhaoNDK18} & \textbf{44.51} & \textbf{29.07} & {21.06} & 15.82 & - & - & - & 16.02 \\
\midrule
answer-focused model \cite{DBLP:conf/emnlp/SunLLHMW18} & \ 42.10 & 27.52 & 20.14 & 15.36 & 46.59 & 33.46 & 24.57 & 18.73 \\
position-aware model \cite{DBLP:conf/emnlp/SunLLHMW18} & \ 42.16 & 27.37 & 20.00 & 15.23 & 47.16 & 34.20 & 24.40 & 18.19 \\
hybrid model \cite{DBLP:conf/emnlp/SunLLHMW18} & {43.02} & {28.14} & {20.51} & {15.64} & 48.24 & 35.95 & 25.79 & 19.45 \\
\midrule
\midrule
\multicolumn{9}{l}{\textbf{Our Model}} \\
\midrule
pointer generator with features (baseline) & 41.25 & 26.76 & 19.53 & 14.89 & 54.04 & 36.68 & 26.62 & 20.15 \\
w/ features + language modeling & 42.80 & 28.43 & \textbf{21.08} & \textbf{16.23}  & {54.47} & {37.30} & \textbf{27.31} & \textbf{20.88} \\
w/o features + language modeling & 42.72 & 27.73 & 20.26 & 15.43 & \textbf{54.62} & \textbf{37.37} & 27.18 & 20.71 \\
w/ features + 1-layer encoder & 42.12 & 27.48 & 20.12 & 15.33 & 53.51 & 36.42 & 26.49 & 20.11 \\
\bottomrule
\end{tabular}}
\caption{Experimental results of our model in different settings comparing with previous methods on two datasets.}
\label{tab:results}
\end{table*}

\section {Model Description}
\begin{figure}
\centering
\includegraphics[scale=0.280]{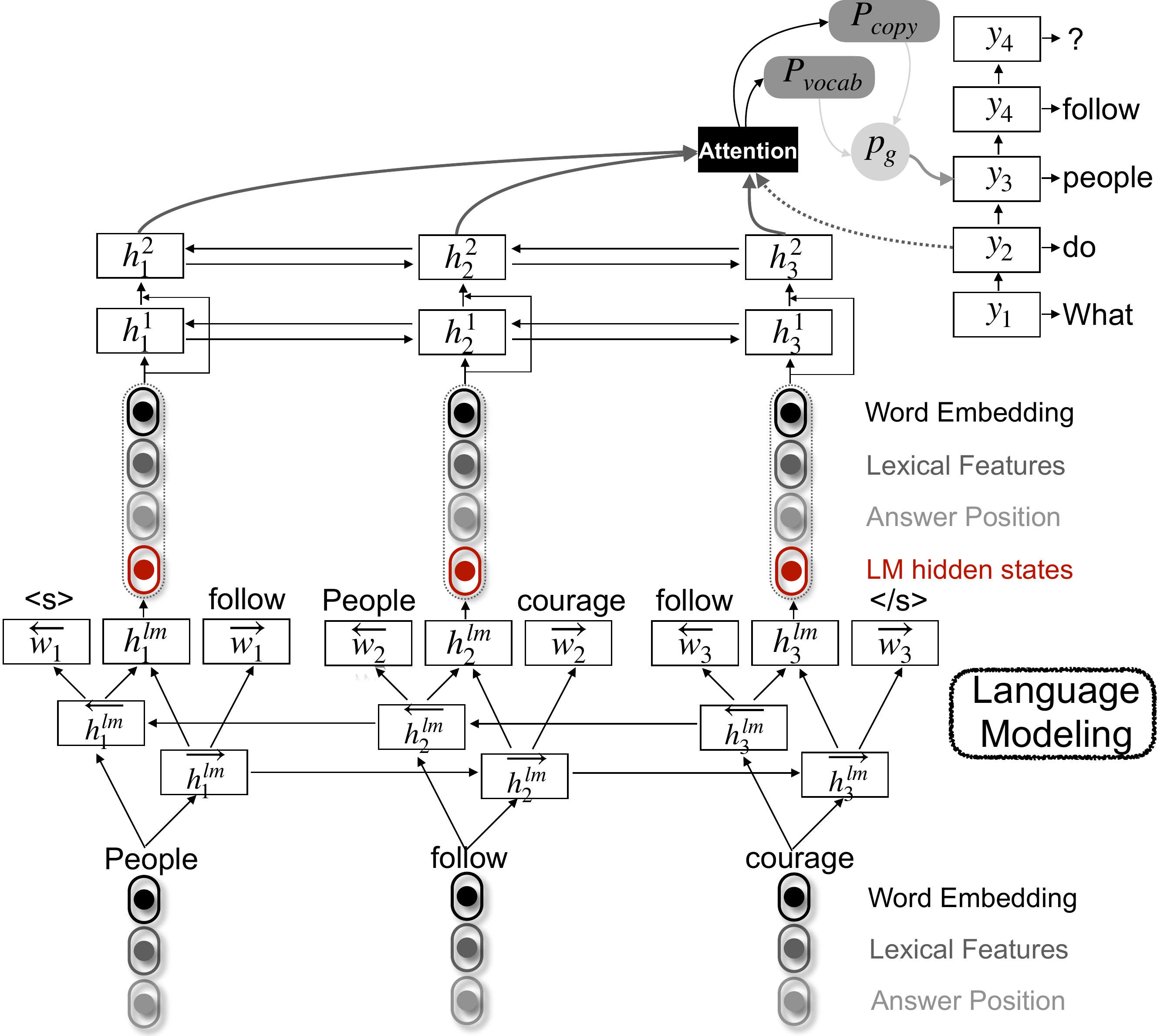}
\caption{\label{joint_model} Overall structure of our joint-learning model.}
\end{figure}

The baseline model is an attention-based seq2seq pointer-generator reinforced by lexical features, like the work of \citet{DBLP:conf/emnlp/SunLLHMW18}. In our proposed model, we employ multi-task learning with language modeling as an auxiliary task for QG. The whole structure of our model is shown in Figure \ref{joint_model}.

\subsection {Feature-enriched Pointer Generator}
The feature-rich encoder is a bidirectional LSTM used to produce a sequence of hidden states $h^L_t$. The encoder takes a sequence of word-and-feature vectors as input ($(x_1, ...,x_T)$), which concatenates the word embedding $e_t$, answer position embedding $a_t$ and lexical feature embedding $l_t$ ($x_t = [e_t;a_t;l_t]$).The lexical feature is composed of POS tags, NER tags, and word case. 

The attention-based decoder is another unidirectional LSTM, which is conditioned on the previous decoder state $s_{i-1}$, decoded word $w_{i-1}$, and context vector $c_{i-1}$ which is generated via attention mechanism \cite{DBLP:journals/corr/BahdanauCB14}:

\begin{align}
\label{decoder}
s_i = LSTM([w_{i-1};c_{i-1}], s_{i-1})
\end{align}
Further, a two-layer feed-forward network is used to produce the vocabulary distribution $P_{vocab}$.

The pointer generator \cite{DBLP:conf/acl/SeeLM17} incorporates a copy mode $P_{copy}(w)$, which allows copying words from the source text via pointing. The final probability distribution is to combine both modes with a generation probability $p_{g} {\in} [0,1]$:
\begin{align}
P(w) & = p_{g}P_{vocab}(w) + (1-p_{g})P_{copy}(w)
\end{align}

The model is trained to minimize the negative log-likelihood of the target sequence. We denote this loss as $E$.

\subsection{Language Modeling}
The language model is to predict the next word and previous word in the sequence with a forward LSTM and a backward LSTM, respectively. First, we feed the input sequence into a bidirectional LSTM to get the hidden representations $h_t^{lm}$.

Then, these states are fed into a softmax layer to predict the next and the previous word:
\begin{align}
\label{lm softmax}
P^{lm}(w_{t+1}|w_{<t+1}) & = softmax({W_f}\overrightarrow{h_t^{lm}}) \\
P^{lm}(w_{t-1}|w_{>t-1}) & = softmax({W_b}\overleftarrow{h_t^{lm}})
\end{align}

The training objective is to minimize the loss function which is defined as the average of the negative log-likelihood of the next word and the previous word in the sequence:
\begin{align}
\label{lm loss}
E^{lm} & = -\frac{1}{T-1}\sum\limits_{t=1}^{T-1}log(P^{lm}(w_{t+1}|w_{<t+1})) \notag \\
& - \frac{1}{T-1}\sum\limits_{t=2}^{T}log(P^{lm}(w_{t-1}|w_{>t-1}))
\end{align}

\subsection{Multi-task Learning}
Instead of sharing representations between two tasks \cite{DBLP:conf/acl/Rei17} or encoding two tasks at the same level \cite{DBLP:conf/emnlp/LiuHSFLH18,DBLP:conf/emnlp/ChenHCL18,DBLP:conf/cvpr/KendallGC18}, we adopt a hierarchical structure to combine the two tasks, by treating language modeling as a low-level task and pointer generator network as high-level, because language modeling is fundamental and its semantic information will benefit question generation. In details, we first feed the input sequence into the language modeling layer to get a sequence of hidden states. Then we concatenate them with the input sequence to obtain the input of the feature-rich encoder.

Finally, the loss of LM is added to the main loss to form a combined training objective:
\begin{align}
E^{total} = E + {\beta}E^{lm}
\end{align}
where $\beta$ is a hyper-parameter, which is used to control the relative importance of two tasks.

\section{Experiments}

\subsection{Dataset}
We conduct experiments on two reading comprehension datasets: SQuAD and MARCO, using the data shared by \citet{DBLP:conf/nlpcc/ZhouYWTBZ17} and \citet{DBLP:conf/emnlp/SunLLHMW18}, where the lexical features are extracted with Stanford CoreNLP. In details, there are 86,635, 8,965 and 8,964 sentence-answer-question triples in the training, development and test set for SQuAD, and 74,097, 4,539 and 4,539 sentence-answer-question triples in the training, development and test set for MARCO.

\subsection{Experiment Settings}
Our vocabulary contains the most frequent 20,000 words in each training set. Word embeddings are initialized with the pre-trained 300-dimensional Glove vectors, and are allowed to be fine-tuned during training. The representations of answer position, POS tags, NER tags and word cases are randomly initialized as 32-dimensional vectors, respectively. The encoder of our baseline model consists of 2 BiLSTM layers, and the hidden size of both the encoder and decoder is set to 512. 

\begin{figure}[!h]
\centering
\includegraphics[scale=0.200]{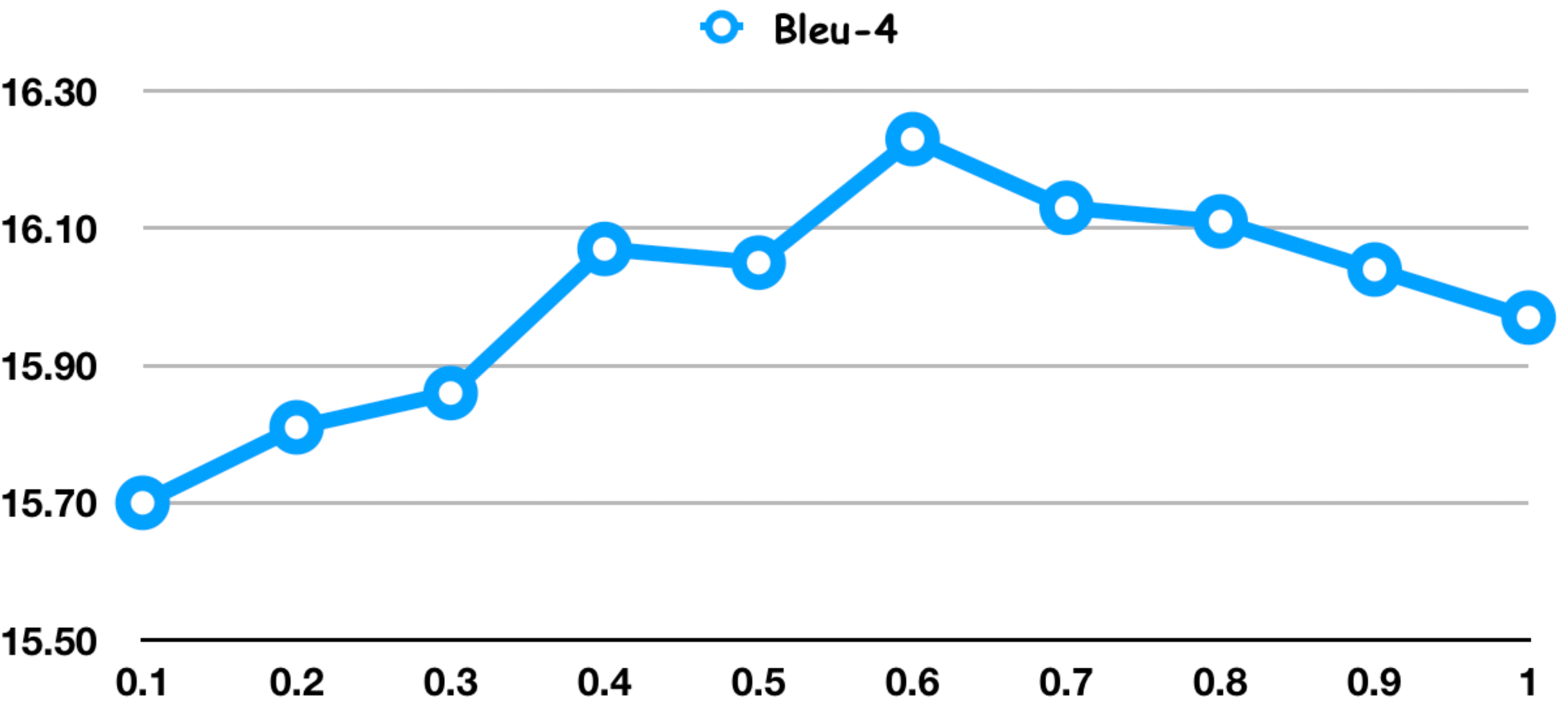}
\caption{\label{beta rate} The impact of $\beta$ on BLEU-4}
\end{figure}

In our joint model, grid search is used to determine $\beta$ and results are shown in Figure \ref{beta rate}. Consequently, we set the value of $\beta$ to 0.6. 

We search the best-trained checkpoint base on the dev-set. In order to mitigate the fluctuation of the training procedure, we then average the nearest 5 checkpoints to obtain a single averaged model. Beam search is used with a beam size of 12.

\begin{table*}[!ht]
\centering
\small
\resizebox{\textwidth}{!}{
\begin{tabular}{l@{\
}|c@{\quad}c@{\quad}c@{\quad}c@{\quad}c@{\quad}c@{\quad}}
\toprule
\textbf{Dataset} & \multicolumn{3}{c}{\textbf{SQuAD}} & \multicolumn{3}{c}{\textbf{MARCO}} \\
\textbf{Model} & \textbf{perplexity} & \textbf{distinct-1} & \textbf{distinct-2} & \textbf{perplexity} & \textbf{distinct-1} & \textbf{distinct-2} \\
\midrule
\midrule
pointer generator with features (baseline) & 41.24 & 9.67 & 39.46 & 17.69 & 16.84 & 45.61 \\
w/ features + language modeling & \textbf{34.09} & \textbf{9.80} & \textbf{40.94} & 15.17 & \textbf{17.31} & \textbf{47.51} \\
w/o features + language modeling & 38.70 & 9.73 & 40.89 & \textbf{14.07} & 17.13 & 46.98 \\
w/ features + 1-layer encoder & 38.40 & 9.56 & 39.71 & 17.61 & 17.10 & 46.87 \\
\bottomrule
\end{tabular}}
\caption{\emph{Perplexity} and \emph{distinct} of different setting models on two datasets}
\label{tab:perplexity}
\end{table*}

\begin{table*}[!ht]
\begin{minipage}[b]{\linewidth}
\centering
\small
\resizebox{\textwidth}{!}{
\begin{tabular}{l@{\
}|c@{\quad}c@{\quad}c@{\quad}c@{\quad}c@{\quad}c@{\quad}}
\toprule
\textbf{Dataset} & \multicolumn{3}{c}{\textbf{SQuAD}} & \multicolumn{3}{c}{\textbf{MARCO}} \\
\textbf{Model} & \textbf{Matching} & \textbf{Fluency} & \textbf{Relevance} & \textbf{Matching} & \textbf{Fluency} & \textbf{Relevance} \\
\midrule
\midrule
pointer generator with features (baseline) & 0.983 & 1.573 & 1.540 & 1.133 & 1.667 & 1.593 \\
+ language modeling & \textbf{1.147} & \textbf{1.690} & \textbf{1.600} & \textbf{1.160} & \textbf{1.720} & \textbf{1.603} \\
\midrule
kendall correlation coefficient & 0.820 & 0.814 & 0.796 & 0.852 & 0.792 & 0.824 \\
\bottomrule
\end{tabular}}
\captionof{table}{Human evaluation results on two datasets.}
\label{tab:human_eval}
\end{minipage}%
\end{table*}

\subsection{Automatic Evaluation}

\textbf{Results on BLEU} The experimental results on BLEU \cite{DBLP:conf/acl/PapineniRWZ02} are illustrated in Table \ref{tab:results}. Our full model (\emph{w/ features + language modeling}) significantly outperforms previous models and achieves state-of-the-art results on both datasets, with 16.23 BLEU-4 score on SQuAD and 20.88 on MARCO respectively. 

\noindent \textbf{Results without Features} To investigate the robustness of our model, we conduct an experiment whose input sequence only takes word embeddings and answer position, but without lexical features (\emph{w/o features + language modeling}). We can see that the auxiliary task of language modeling boosts model performance on both datasets, demonstrating that our model guarantees higher stability because it does not depend on the quality of lexical features. Therefore, our model can apply to low-resource languages where there is not adequate data for training a well-performed model for lexical features extraction. 

\noindent \textbf{Results with a 3-layer Encoder} To validate that we gain the improvement not due to a deeper network, we replace the language modeling module with one encoder layer, that is to say, we adopt a 3-layer encoder. Comparing this model (\emph{w/ features+1-layer encoder}) with the full model (\emph{w/ features+language modeling}), we can see that our joint-learning model performs better than simply adding an extra encoding layer. The results on MARCO also clearly show that a deeper network does not guarantee better performance.

\noindent \textbf{Perplexity and Diversity}
Since BLEU only measures a hard matching between references and generated text, we further adopt \emph{perplexity} and \emph{distinct} \cite{DBLP:conf/naacl/LiGBGD16} to judge the quality of generated questions. The results in Table \ref{tab:perplexity} indicate that the language modeling task helps the model to generate more fluent and readable questions. Besides, the generated questions have better diversity.

\subsection{Human Evaluation}
For a better study on the quality of generations, we perform human evaluation. Three annotators are asked to grade the generated questions in three aspects: $matching$ indicates whether a question can be answered with the given answer; $fluency$ indicates whether a question is fluent and grammatical; $relevance$ indicates whether a question can be answered according to the given context. The rating score ranges from 0 to 2. We randomly sample 100 cases from each dataset for evaluation. Results are displayed in Table \ref{tab:human_eval}. The coefficient between human judges is high, validating a high quality of our annotation. The results show that by incorporating language modeling, the generated questions receive higher scores across all three metrics. 

\begin{table}[ht]
    \centering
    \small
    \begin{tabular}{p{7.25cm}}
    \toprule
        \textbf{Context:} Prior to the early 1960s, access to the forest's interior was highly restricted, and the forest remained basically intact. \\
        \textbf{Answer:} The early 1960s \\
        \textbf{Reference:} Accessing the Amazon rainforest was restricted before what era?\\
        \textbf{Baseline:} When did access to the forest's interior? \\
        \textbf{Joint-model:} When did access to the forest's interior become restricted? \\
        \midrule
        \textbf{Context:} This teaching by Luther was clearly expressed in his 1525 publication on the bondage of the will, which was written in response to on free will by Desiderius Erasmus (1524). \\
        \textbf{Answer:} 1525 \\
        \textbf{Reference:} When did Luther publish on the bondage of the will?\\
        \textbf{Baseline:} In what year was the bondage of the will on the bondage of the will? \\
        \textbf{Joint-model:} When was the bondage of the will published? \\
    \bottomrule
    \end{tabular}
    \caption{Examples of generated questions by different models.}
    \label{tab:my_label}
\end{table}

\subsection{Case Study}
Further, Table \ref{tab:my_label} gives two examples of the generated questions on SQuAD dataset, by the baseline model and our joint model respectively. It is obvious that questions generated by our proposed model are more complete and grammatical.

\section{Conclusion}
This paper proves that equipped with language modeling as an auxiliary task, the neural model for QG can learn better representations that help the decoder to generate more accurate and fluent questions. In future work, we will adopt the auxiliary language modeling task to other neural generation systems to test its generalization ability. 

\section*{Acknowledgments}
We thank Weikang Li and Xin Jia for their valuable comments and suggestions. This work is supported by the National Natural Science Foundation of China (61773026, 61572245).

\bibliography{emnlp-ijcnlp-2019}
\bibliographystyle{acl_natbib}

\appendix

\end{document}